\documentclass[letterpaper]{article} 
\usepackage{aaai24}  
\usepackage{times}  
\usepackage{helvet}  
\usepackage{courier}  
\usepackage[hyphens]{url}  
\usepackage{graphicx} 
\usepackage{natbib}  
\usepackage{caption} 
\usepackage{algorithm}
\usepackage{algorithmic}

\usepackage{newfloat}
\usepackage{listings}
\DeclareCaptionStyle{ruled}{labelfont=normalfont,labelsep=colon,strut=off} 
\floatstyle{ruled}
\newfloat{listing}{tb}{lst}{}
\floatname{listing}{Listing}
\usepackage{amsmath}
\usepackage{amssymb}
\usepackage{subfigure}
\usepackage{bibentry}

\title{Reliable Conflictive Multi-View Learning}
\author{
    Cai Xu,
    Jiajun Si,
    Ziyu Guan, 
    Wei Zhao\thanks{Corresponding author}, 
    Yue Wu, 
    Xiyue Gao
}
\affiliations{
    School of Computer Science and Technology, Xidian University, China\\
    \{cxu@, jiajunsi@stu., zyguan@, ywzhao@mail., ywu@, xygao@\}xidian.edu.cn
}

\usepackage{bibentry}

\begin{document}

\maketitle

\begin{abstract}
Multi-view learning aims to combine multiple features to achieve more comprehensive descriptions of data. Most previous works assume that multiple views are strictly aligned. However, real-world multi-view data may contain low-quality conflictive instances, which show conflictive information in different views. Previous methods for this problem mainly focus on eliminating the conflictive data instances by removing them or replacing conflictive views. Nevertheless, real-world applications usually require making decisions for conflictive instances rather than only eliminating them. To solve this, we point out a new Reliable Conflictive Multi-view Learning (RCML) problem, which requires the model to provide decision results and attached reliabilities for conflictive multi-view data. We develop an Evidential Conflictive Multi-view Learning (ECML) method for this problem. ECML first learns view-specific evidence, which could be termed as the amount of support to each category collected from data. Then, we can construct view-specific opinions consisting of decision results and reliability. In the multi-view fusion stage, we propose a conflictive opinion aggregation strategy and theoretically prove this strategy can exactly model the relation of multi-view common and view-specific reliabilities. Experiments performed on 6 datasets verify the effectiveness of ECML. The code is released at https://github.com/jiajunsi/RCML.
\end{abstract}

\section{Introduction}
Artificial intelligence systems usually perceive and understand the world from multi-view data. For example, automated vehicle systems sense their surroundings through multiple sensors (e.g., camera, lidar, radar); recommender systems capture users' preferences from their multi-view generated content such as textual review and visual review. Integrating the consistent and complementary information of multiple views could obtain a more comprehensive description of data instances, which boosts various tasks such as clustering \cite{xu2019adversarial,NEURIPS2022_a6610efd,wen2022survey,ektefaie2023multimodal}, retrieval \cite{mostafazadeh2017image,10.1145/3503161.3547922} and recommendation \cite{10.1145/3551349.3559496,tan20224sdrug}.

\begin{figure}[t]
\centering
\includegraphics[width=0.9\columnwidth]{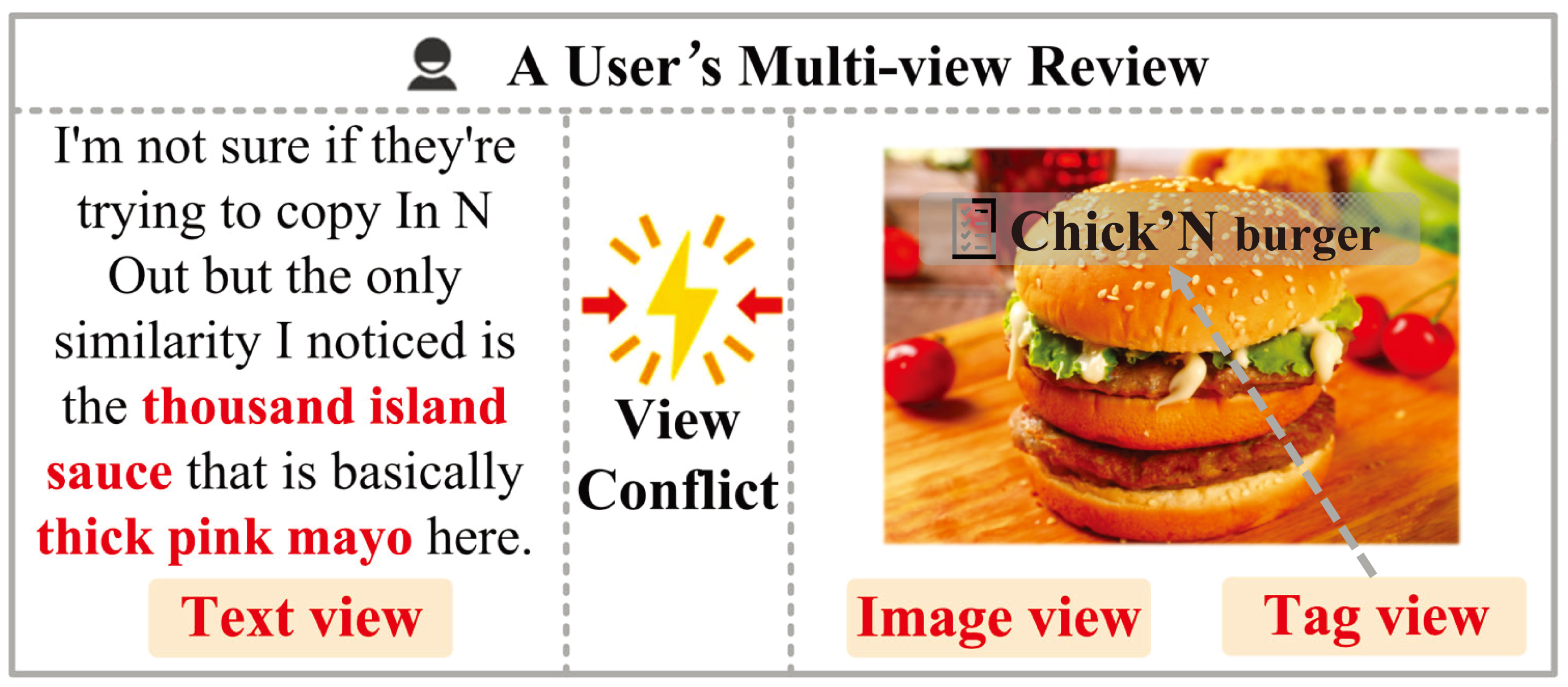}
\caption{Visualization of the conflictive multi-view data: the text view is related to food ``sauce"; however, the other views show conflictive information, i.e., food ``burger".}
\label{fig:motivation}
\end{figure}

Most of the previous studies on multi-view learning \cite{liang2022foundations,zhao2023tensorized,zhang2023multi} always assume that data of different views are strictly aligned. For example, different views consistently belong to the ground-truth category in a classification task. However, in real-world settings, this assumption cannot always be guaranteed. Fig. \ref{fig:motivation} visualizes a case of users' multi-view generated content: the text and image views show conflictive food categories. As a result, this conflictive information in different views makes most multi-view learning methods inevitably degenerate or even fail. 

The prevalent solutions to this problem mainly aim to eliminate the conflictive data instance. Pioneer works regard the conflictive data as outliers \cite{10.1145/2505515.2507840,9122431}. They usually consist of 3 steps: 1) measuring the consistency of views; 2) identifying outliers as instances with significant inconsistencies across views; 3) removing outliers to construct a clean dataset. Recently, some multi-view learning methods \cite{NEURIPS2020_1e591403,10.1007/978-3-031-33374-3_18} dedicate to learning alignment relations for the original data and constructing new data instances accordingly. For example, the text view in Fig. \ref{fig:motivation} would be replaced with this view of another aligned instance. Therefore, the conflict in the original instances would be solved. 

Nevertheless, real-world applications usually require making decisions for conflictive instances rather than only eliminating them. For instance, recommender systems need to predict users' preferences from their conflictive multi-view review. Considering the decision of a conflictive instance might be unreliable, we need the model can answer ``should the decision be reliable?". Therefore, we point out a new problem in this work, Reliable Conflictive Multi-view Learning (RCML) problem, which requires the model to provide the decision results and the attached reliabilities for conflictive multi-view data.

\begin{figure*}[t]
    \centering
    \includegraphics[width=0.9\textwidth]{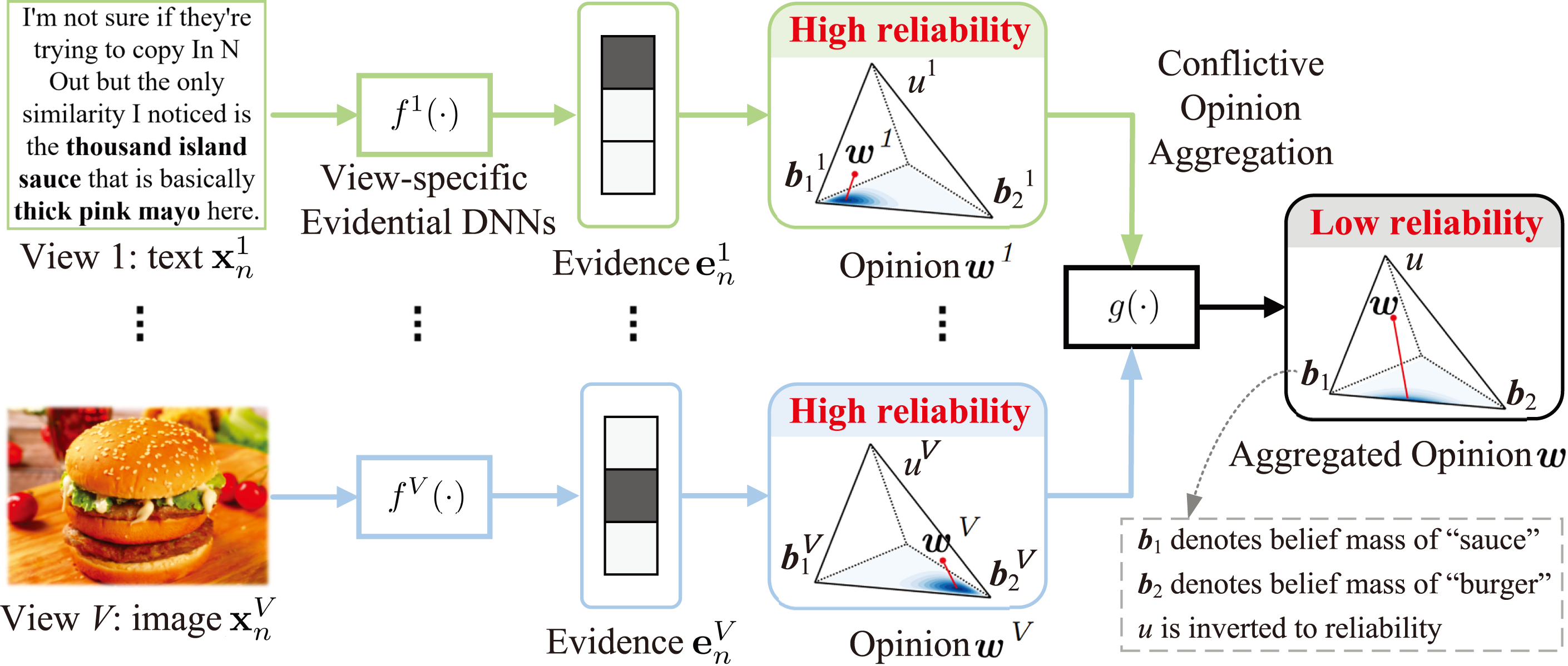}
    \caption{ Illustration of ECML. View-specific DNNs collect evidence, which could be termed as the amount of support to each category. Then we form view-specific opinions consisting of belief masses of all categories and uncertainty (inverted to reliability). Finally, we integrate opinions by conflictive opinion aggregation. The uncertainty of the aggregated opinion might increase if view-specific opinions are conflictive.}
    \label{fig:model}
\end{figure*}

In this paper, we propose an Evidential Conflictive Multi-view Learning (ECML) method for the RCML problem. As shown in Fig. \ref{fig:model}, we first construct the view-specific evidential Deep Neural Networks (DNNs) to learn view-specific evidence, which could be termed as the amount of support to each category collected from data. Then the view-specific distributions of the class probabilities are modeled by Dirichlet distribution, parameterized with view-specific evidence. From the distributions, we can construct opinions consisting of belief mass vector and decision reliability. Specifically, we calculate the conflictive degree according to the projected distance and conjunctive certainty among views. In the multi-view fusion stage, we propose a conflictive opinion aggregation strategy and establish a simple and effective average pooling fusion layer accordingly. We theoretically prove the final  reliability would be less than view-specific reliabilities for conflictive instances.

The first contribution is recognizing the importance of explicitly providing decision results and associated reliabilities when dealing with conflicting multi-view data. Another contribution is that we develop a conflictive opinion aggregation strategy and theoretically prove it can exactly model the relation of multi-view common and view-specific reliabilities. Finally, we empirically compare ECML with state-of-the-art multi-view learning baselines on 6 publicly available datasets. Experiment results show that ECML outperforms baseline methods on accuracy, reliability and robustness.

\section{Related Work}

We briefly review related work about conflictive multi-view learning and uncertainty-aware deep learning.

\textbf{Conflictive Multi-View Learning:} 
The prevalent conflictive multi-view learning methods are mainly dedicated to eliminating the conflictive data instance. One line is based on multi-view outlier detection, which is developed to detect outliers with abnormal behavior in the multi-view context. There are two main categories for classifying these methods: cluster-based \cite{Huang_Ren_Pu_Huang_Xu_He_2023,zhang2023let} and self-representation-based \cite{wang2019adversarial,wen2023highly}. Cluster-based methods employ separate clustering in each view and generate affinity vectors for each instance accordingly \cite{10.1145/2505515.2507840, 8047342}. Outliers are subsequently identified by comparing the affinity vectors across multiple views. Self-representation-based methods identify outliers by recognizing that they are difficult to represent using normal views \cite{9122431}.

Another line is based on partially view-aligned multi-view learning \cite{wen2023unpaired,zhang2021late}. Earlier work \cite{10.1007/978-3-642-15552-9_41} introduces weakly-paired maximum covariance analysis to overcome the limitations of unaligned data. Recently, Huang et al. \cite{NEURIPS2020_1e591403} employ the differentiable agent of the Hungarian algorithm to establish alignment relationships for unaligned data. Along this line, researchers propose noise-robust contrastive learning \cite{Yang_2021_CVPR}, the self-focused mechanism \cite{10.1007/978-3-031-33374-3_18}, et al., to compute the alignment matrix. However, these methods aim to eliminate conflictive instances, while real-world applications usually require making decisions for them. Therefore, we propose to make reliable decisions for conflictive instances.

\textbf{Uncertainty-aware Deep Learning:} Deep neural networks have achieved remarkable success in various tasks, but often fail to capture the uncertainty of their predictions, especially for low-quality data \cite{wen2023deep}. Uncertainty can be categorized into aleatoric uncertainty (related to data uncertainty) and epistemic uncertainty (associated with model uncertainty). Deep learning for estimating uncertainty \cite{gawlikowski2023survey} can be classified into: single deterministic method \cite{NEURIPS2018_a981f2b7}, bayesian neural networks \cite{pmlr-v48-gal16}, ensemble methods \cite{NIPS2017_9ef2ed4b} and test-time augmentation methods \cite{pmlr-v124-lyzhov20a}. Specifically, the representative single deterministic method, Evidential Deep Learning (EDL) \cite{NEURIPS2018_a981f2b7} calculates the category-specific evidence according to a single DNN. 

Recently, researchers extend EDL to the multi-view learning area. The pioneering work, Trusted Multi-View Classification (TMC) \cite{han2021trusted} involves Dempster’s combination rule, which assigns small weights to highly uncertain views. Following this line, multiple opinion aggregation methods \cite{jung2022uncertainty,Liu_Yue_Chen_Denoeux_2022,liu2023safe,zhang2023provable} are proposed. However, an important characteristic of them is ``After integrating another opinion into the original opinion, the obtained uncertainty mass will be reduced" \cite{9767662}. We argue that this since when incorporating an unreliable or conflicting opinion, the uncertainty should increase. To solve this, we propose a conflictive opinion aggregation strategy and theoretically prove the uncertainty would increase for conflictive instances.

\section{The Method}
In this section, we first define the RCML problem, then present ECML in detail, together with the theoretical prove discussion, and analyses.

\begin{figure}[t]
\centering
\includegraphics[width=0.95\columnwidth]{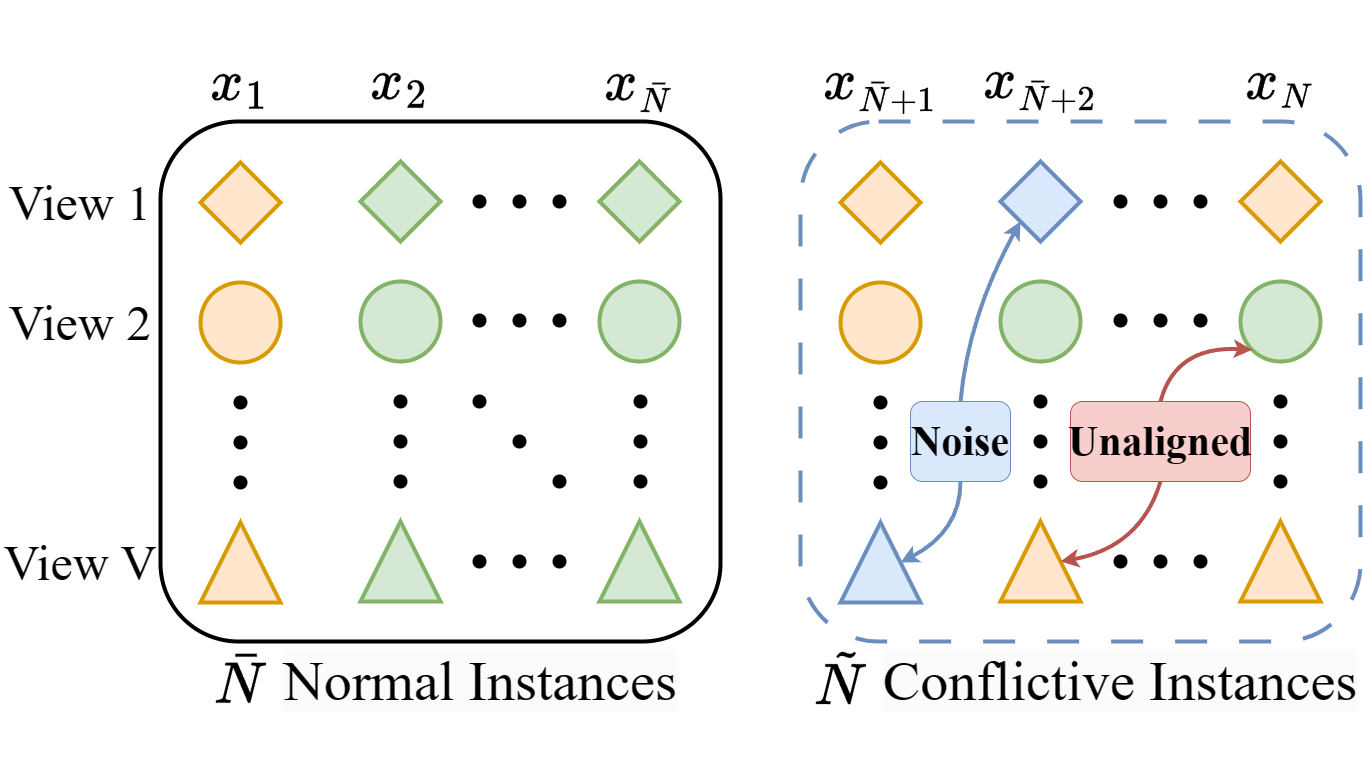}
\caption{Notations for conflictive multi-view data. The two categories are marked as yellow and green respectively. Conflictive instances contain noise and unalignment views: noise views are marked as blue and do not belong to any ground-truth categories; unaligned views show different categories from other views. }
\label{fig:data}
\end{figure}

\subsection{Problem Definition}
In the RCML setting, suppose we are given a dataset with $V$ views, $\bar{N}$ normal instances and $\tilde{N}$ conflictive instances as shown in Fig. \ref{fig:data}. We use $\mathbf{x}_{n}^{v} \in \mathbb{R}^{D_{v} }(v = 1,...,V)$ to denote the feature vector for the $v$-th view of the $n$-th instance $(n=1,...,N)$, where $D_{v}$ is the dimensionality of the $v$-th view. The one-hot vector $\mathbf{y}_{n} \in \left \{ 0, 1 \right \}^K$ denotes the ground-truth label of the $n$-th instance, where $K$ is the total of all categories. The training tuples $ \{ \left \{ \mathbf{x}_{n}^{v}  \right \}_{v=1}^{V}, \mathbf{y}_{n} \} _{n=1}^{\bar{N}_{train}}$ contain $\bar{N}_{train}$ normal instances. The other $\bar{N} - \bar{N}_{train}$ normal instances and  $\tilde{N}$ conflictive instances form the test set. The goal of RCML is to accurately predict $\mathbf{y}_{n}$ for the test instances and provide the attached prediction uncertainties ${{u_{n} \in [0,1]}}$\footnote{The ``ground truth” uncertainties are usually not available. We manually construct conflictive instances and expect large prediction uncertainties. We elaborate the construction approach in the experiment section.} to measure the decision reliability ($ 1 - u_{n} $). 

\subsection{Evidential Conflictive Multi-view Learning}

As shown in Fig. \ref{fig:model}, the overall architecture consists of view-specific evidential learning and evidential multi-view fusion stages. In the first stage, we learn view-specific evidence by evidential DNNs, which could be termed as the amount of support to each category collected from data. Then the view-specific distributions of the class probabilities are modeled by Dirichlet distribution, parameterized with view-specific evidence. From the distributions, we can construct opinions consisting of belief mass vector and decision reliability. Specifically, we calculate the conflictive degree according to the projected distance and conjunctive certainty among views. By minimizing the conflictive degree, we force the view-specific DNNs to well capture multi-view common information in the training stage. This would reduce the decision conflict caused by view-specific DNNs, i.e., normal instances mistake for conflictive instance since view-specific models make wrong decisions. In the second stage, we propose a conflictive opinion aggregation strategy and establish a simple and effective average pooling fusion layer accordingly. Details will be elaborated as below.

\subsubsection{View-specific Evidencial Deep Learning.}

Most existing deep multi-view learning methods commonly rely on employing a softmax layer atop deep DNNs for classification purposes. However, these softmax-based DNNs face limitations in accurately estimating predictive uncertainty. This is due to the fact that the softmax score essentially provides a single-point estimation of a predictive distribution, leading to over-confident outputs even in cases of false predictions.

To solve this, we employ EDL \cite{NEURIPS2018_a981f2b7} in the view-specific learning stage. EDL was developed to address the above limitation by introducing the evidence framework of subjective logic (SL) \cite{josang2016subjective}. In this context, evidence refers to the metrics collected from the input to support the classification process. We collect evidence, $\{ \boldsymbol{\mathit{e}}_{n}^{v}  \}$ by view-specific evidential DNNs $\{ f^v (\cdot) \}_{v=1}^V$.

For $K$ classification problems, a multinomial opinion over a specific view of an instance ($\mathbf{x}_{n}^{v}$)\footnote{In this section, we omit the super- and sub-scripts for clarify.} can be represented as an ordered triplet $\mathbf{\mathit{w}} = (\boldsymbol{\mathit{b}}, \mathit{u}, \boldsymbol{\mathit{a}})$, where the belief mass $\boldsymbol{\mathit{b}} = (\mathit{b}_{1},...,\mathit{b}_{k})^{\top}$ assigns belief masses to possible values of the instance based on the evidence support for each value. The uncertainty mass $\mathit{u}$ captures the degree of ambiguity or vacuity in the evidence, while the base rate distribution $\boldsymbol{\mathit{a}} = (\mathit{a}_{1},...,\mathit{a}_{k})^{\top}$ represents the prior probability distribution over each class $k$. Subjective logic dictates that both $\boldsymbol{\mathit{b}}$ and $\mathit{u}$ must be non-negative and their sum should equal one:
\begin{align}
\sum_{k=1}^{K} \mathit{b}_{k} + \mathit{u} = 1, \forall k\in \left [ 1,...,K \right ],
\end{align}
where $\mathit{b}_{k} \ge 0$ and $\mathit{u} \ge 0$. The projected probability distribution of multinomial opinions is given by:
\begin{align}
\mathit{P}_{\mathit{k}} = \mathit{b}_{\mathit{k}} + \mathit{a}_{\mathit{k}}\mathit{u}, \forall k\in \left [ 1,...,K \right ].
\end{align}

Normally, the prior probabilities are manually set according to prior knowledge. For example, a common approach is to set the prior probabilities equal for each category, denoted as $\mathit{a}_{\mathit{k}} = 1/K$. This implies all categories have similar data instance numbers. 

The Dirichlet Probability Density Function (PDF) is used for forming category distribution. It can model second-order uncertainty, while the probability values in the softmax layer only capture first-order uncertainty. The probability density function of the Dirichlet distribution is given by:
\begin{align}
\mathit{D}(\mathbf{p}|\boldsymbol{\alpha} ) =  \left\{
\begin{matrix}
\frac{1}{\mathit{B}(\boldsymbol{\alpha})}  {\textstyle \prod_{k=1}^{K}\mathit{p}_{k}^{\mathit{\alpha}_{k}-1} }, &for \ \mathbf{p}\in \mathcal{S}_{K},
\\0, &otherwise,
\end{matrix}
\right.
\end{align}
where  $\boldsymbol{\mathit{p}} = (\mathit{p}_{1},...,\mathit{p}_{k} )^{\top }$ is the probability that the instance is assigned to $k$-th class,  $\boldsymbol{\mathit{\alpha}} = (\mathit{\alpha}_{1},...,\mathit{\alpha}_{k} )^{\top }$ represents the Dirichlet parameters, $\mathcal{S}_{K}$ is the $K$-dimensional unit simplex, defined as:
\begin{align}
\mathcal{S}_{K}=\left \{ \mathbf{p}|\sum_{k=1}^{K}\mathit{p}_{k}=1 \ and \ 0 \le \mathit{p}_{1},...,\mathit{p}_{k} \le 1   \right \}, 
\end{align}
and $\mathit{B}(\boldsymbol{\alpha})$ is the $K$-dimensional multinomial beta function.

The Dirichlet PDF naturally reflects a random sampling of statistical events, which is the basis for the aleatory interpretation of opinions as statistical measures of likelihood. Then, uncertainty mass can be well expressed in the form of Dirichlet PDFs. Uncertainty mass in the Dirichlet model reflects the vacuity of evidence. Interpreting uncertainty mass as vacuity of evidence reflects the property that “the fewer observations the more uncertainty mass”.

We calculate the Dirichlet distribution parameters $\boldsymbol{\alpha}$ by $\boldsymbol{\alpha} = \boldsymbol{\mathit{e}} + \textbf{1}$ to guarantee the parameters are larger than one, and hence the Dirichlet distribution is non-sparse. A mapping between the multinomial opinion and Dirichlet distribution can be given by:
\begin{align}
\mathit{b_{k}}=\frac{\mathit{e}_{k}}{S}=\frac{\mathit{\alpha_{k}}-1}{S}, \mathit{u}=\frac{K}{S}, 
\end{align}
where $S= {\textstyle \sum_{k=1}^{K}}\left ( \mathit{e}_{k}+1  \right )  = {\textstyle \sum_{k=1}^{K}}\mathit{\alpha }_{k} $ is the Dirichlet strength, $ \boldsymbol{\mathit{e}} = (\mathit{e}_{1},...,\mathit{e}_{K} )^{\top }$. It is important to note that the level of uncertainty is inversely proportional to the amount of total evidence available. In the absence of any evidence, the belief for each view is 0, resulting in maximum uncertainty, i.e., 1. Specifically, the class probability $\mathit{p}_{k}$ could be computed as $\mathit{p}_{k} = \mathit{\alpha }_{k}/{S}$.

Through the view-specific evidential learning stage, we obtain the view-specific opinion and the corresponding category distribution.

\subsubsection{Evidential Multi-view Fusion via Conflictive Opinion Aggregation.}

In this subsection, we focus on multi-view fusion according to view-specific opinions. The noise views of the conflictive multi-view data would show high uncertainty. We would diminish their impact in the fusion stage. The unaligned views of the conflictive multi-view data would provide highly conflicting opinions with low uncertainty, which may indicate that one or more views are unreliable. In this case, we are hard to judge which view is high-quality. In fact, the uncertainty of multi-view learning results should not decrease with the increase of the number of views, but should be related to the quality of the perspectives to be fused, especially when the learning results of two views conflict. To solve this, we propose a new conflictive opinion aggregation method.

\textbf{Definition 1 Conflictive Opinion Aggregation.} Let $\boldsymbol{\mathit{w} } ^{\mathit{A}}= (\boldsymbol{\mathit{b} }^{\mathit{A}}, \mathit{u}^{\mathit{A}}, \boldsymbol{\mathit{a} }^{\mathit{A}})$ and $\boldsymbol{\mathit{w} } ^{\mathit{B}}= (\boldsymbol{\mathit{b} }^{\mathit{B}}, \mathit{u}^{\mathit{B}}, \boldsymbol{\mathit{a} }^{\mathit{B}})$ be the opinions of view $\mathit{A}$ and $\mathit{B}$ over the same instance, respectively. The conflictive aggregated opinion $\boldsymbol{\mathit{w} } ^{A\underline{\Diamond } B}$ is calculated in the following manner:
\begin{gather}
\boldsymbol{\mathit{w} } ^{A\underline{\Diamond } B}= \boldsymbol{\mathit{w} } ^{\mathit{A}} \underline{\Diamond } \boldsymbol{\mathit{w} } ^{\mathit{B}} =  (\boldsymbol{\mathit{b} }^{A\underline{\Diamond } B}, \mathit{u}^{A\underline{\Diamond } B}, \boldsymbol{\mathit{a} }^{A\underline{\Diamond } B}),\\
{\mathit{b}}_{k}^{A\underline{\Diamond } B}=\frac{{\mathit{b}}_{k}^{A}\mathit{u^{B}}+{\mathit{b}}_{k}^{B}\mathit{u^{A}}}{\mathit{u^{A}}+\mathit{u^{B}}},\\
\mathbf{\mathit{u}}^{A\underline{\Diamond } B}=\frac{2\mathit{u^{A}}\mathit{u^{B}}}{\mathit{u^{A}}+\mathit{u^{B}}}, 
\boldsymbol{\mathit{a}}_{k}^{A\underline{\Diamond } B}=\frac{\boldsymbol{\mathit{a}}_{k}^{A}+\boldsymbol{\mathit{a}}_{k}^{B}}{2}.
\end{gather}

The opinion $\boldsymbol{\mathit{w} } ^{A\underline{\Diamond } B}$ represents the combination of the dependent opinions of $A$ and $B$. This combination is achieved by mapping the belief opinions to evidence opinions using a bijective mapping between multinomial opinions and the Dirichlet distribution. Essentially, the combination rule ensures that the quality of the new opinion is proportional to the combined one. In other words, when a highly uncertain opinion is combined, the uncertainty of the new opinion is larger than the original opinion. The averaging belief fusion can be computed simply by averaging the evidence. A more detailed explanation is shown in Proposition 1.

Following Definition 1, we can fusion the final joint opinions $\boldsymbol{\mathit{w} }$ from different views with the following rule:
\begin{align}
\boldsymbol{\mathit{w} } = \boldsymbol{\mathit{w} } ^{\mathit{1}} \underline{\Diamond } \boldsymbol{\mathit{w} } ^{\mathit{2}}
\underline{\Diamond } ...
\underline{\Diamond } \boldsymbol{\mathit{w} } ^{\mathit{V}}.
\end{align}

According to the above fusion rules, we can get the final multi-view joint opinion, and thus get the final probability of each class and the overall uncertainty.

We also aim to: 1) ensure the consistency of the model in different views during the training stage (using normal instances); 2) get an intuitive sense of the level of conflict. Therefore, we introduce a measure named the conflictive degree in Definition 2, which is established according to opinion entropy.

\textbf{Definition 2 Conflictive Degree.} Given two opinions $\boldsymbol{w}^{A}$ and $\boldsymbol{w}^{B}$ over an instance, the conflictive degree between $\boldsymbol{w}^{A}$ and $\boldsymbol{w}^{B}$ is defined as:
\begin{align}
c(\boldsymbol{w}^{A}, \boldsymbol{w}^{B}) = c_p (\boldsymbol{w}^{A}, \boldsymbol{w}^{B}) \cdot c_c (\boldsymbol{w}^{A}, \boldsymbol{w}^{B}),
\end{align}
where $c_p(\boldsymbol{w}^{A}, \boldsymbol{w}^{B})$ is the projected distance between $\boldsymbol{w}^{A}$ and $\boldsymbol{w}^{B}$, $c_c(\boldsymbol{w}^{A}, \boldsymbol{w}^{B})$ is the conjunctive certainty between $\boldsymbol{w}^{A}$ and $\boldsymbol{w}^{B}$, which can be formulated as follows:
\begin{align}
c_p(\boldsymbol{w}^{A}, \boldsymbol{w}^{B})=\frac{\sum_{k=1}^{K}\left | p^{A}_{k}-p^{B}_{k} \right |  }{2},\\
c_c(\boldsymbol{w}^{A}, \boldsymbol{w}^{B})=(1-u^{A})(1-u^{B}).
\end{align}

Intuitively, this metric ensures two things: (1) The scenario where $c = 0$ arises when the same projected probability distributions are observed, indicating non-conflicting opinions; (2) $c = 1$ arises when absolute opinions are present but with different projected probabilities. Specifically, when $c_c = 0$, it indicates that the vacuous condition is present in one or both opinions. On the other hand, when $c_c = 1$, it signifies that the opinions are considered credible, meaning they have zero uncertainty mass. 

\subsubsection{Loss Function.}

In this subsection, we will introduce the training DNN to obtain the multi-view joint opinion. Traditional DNN can be easily converted into evidential DNN with minimal modifications, as demonstrated in \cite{NEURIPS2018_a981f2b7}. This transformation primarily involves replacing the softmax layer with an activation layer (e.g., ReLU) and considering the non-negative output of this layer as evidence. By doing so, we can obtain the parameters of the Dirichlet distribution.

For instance $\{ \mathbf{x}_{n}^v \}_{v=1}^V$, $\mathbf{e}^{v}_{n}=f^{v}(\mathbf{x}^{v}_{n})$ represent the evidence vector predicted by the network for the classification. $\boldsymbol{\alpha}^{v}_{n}=\mathbf{e}^{v}_{n} + \mathbf{1}$ is the parameters of the corresponding Dirichlet distribution. In the case of conventional neural network-based classifiers, the cross-entropy loss is typically employed. However, we need to adapt the cross-entropy loss to account for the evidence-based approach:
\begin{align}
\nonumber \mathit{L_{ace}}(\boldsymbol{\alpha}_{n}) &= \int \left [ \sum_{j=1}^{K}-y_{nj}\log_{}{p_{nj}}\right ] \frac{\textstyle \prod_{j=1}^{K}p_{nj}^{{\alpha}_{nj}-1}}{B\left ( \boldsymbol{\alpha}_{n} \right)}d\mathbf{p}_{n}\\ 
&=\sum_{j=1}^{K}y_{nj}(\psi(S_{n})-\psi({\alpha}_{nj})),
\end{align}

where $\psi(\cdot)$ is the digamma function. 

The above loss function does not guarantee that the evidence generated by the incorrect labels is lower. To address this issue, we can introduce an additional term in the loss function, namely the Kullback-Leibler (KL) divergence:
\begin{align}
L_{KL}(\boldsymbol{\alpha}_{n}) 
&= KL \left[D(\boldsymbol{p}_{n}|\tilde{\boldsymbol{\alpha}}_{n})\parallel D(\boldsymbol{p}_{n}|\mathbf{1})\right] \\ \nonumber
&= \log_{}{(\frac{\Gamma( {\textstyle \sum_{k=1}^{K}}\tilde{\alpha}_{nk})}{\Gamma(K) {\textstyle \prod_{k=1}^{K}\Gamma(\tilde{\alpha}_{nk})} })} \\ \nonumber
&+ \sum_{k=1}^{K}(\tilde{\alpha}_{nk}-1)\left [ \psi(\tilde{\alpha}_{nk})-\psi (\sum_{j=1}^{K}\tilde{\alpha}_{nj}) \right ],
\end{align}
where $D(\mathbf{p}_{n}|\mathbf{1})$ is the uniform Dirichlet distribution, $\tilde{\boldsymbol{\alpha}}_{n} = \mathbf{y}_{n} + (\mathbf{1}-\mathbf{y}_{n}) \odot \boldsymbol{\alpha}_{n}$ is the Dirichlet parameters after removal of the non-misleading evidence from predicted parameters $\boldsymbol{\alpha}_{n}$ for the $n$-th instance, and $\Gamma(\cdot)$ is the gamma function. 

Therefore, given the Dirichlet distribution with parameter $\alpha_{n}$ for the $n$-th instance, the loss is:
\begin{align}
L_{acc}(\boldsymbol{\alpha}_{n}) = L_{ace}(\boldsymbol{\alpha}_{n}) \ + \ \lambda_{t}L_{KL}(\boldsymbol{\alpha}_{n}),
\end{align}
where $\lambda_{t} = min(1.0, t/T) \in \left [ 0, 1 \right] $ is the annealing coefficient, $t$ is the index of the current training epoch, and $T$ is the annealing step. By gradually increasing the influence of KL divergence in loss, premature convergence of misclassified instances to uniform distribution can be avoided.

In order to ensure the consistency of results between different opinions during training, minimizing the degree of conflict between opinions was adopted. The consistency loss for the instance $\{ \mathbf{x}_{n}^v \}_{v=1}^V$ is calculated as:
\begin{align}
L_{con} = \frac{1}{V-1} \sum_{p=1}^{V}\left(\sum_{q \ne p}^{V} c(\boldsymbol{w}^{p}_{n}, \boldsymbol{w}^{q}_{n}) \right).
\end{align}

To sum up, the overall loss function for a specific instance $\{ \mathbf{x}_{n}^v \}_{v=1}^V$ can be calculated as:
\begin{align}
L =  L_{acc}(\boldsymbol{\alpha}_{n}) + \beta \sum_{v=1}^{V} L_{acc}(\boldsymbol{\alpha}_{n}^{v}) + \gamma L_{con}.
\end{align}

The model optimization is elaborated in Algorithm 1 (Technical Appendix).

\subsection{Discussion and Analyses}

In this subsection, we theoretically analyze the advantages of ECML, especially the conflictive opinion aggregation for the conflictive multi-view data. The following propositions provide the theoretical analysis to support the conclusions. The proof is shown in the Technical Appendix.

\textbf{Proposition 1} The conflictive opinion aggregation $\boldsymbol{\mathit{w} } ^{A\underline{\Diamond } B}= \boldsymbol{\mathit{w} } ^{\mathit{A}} \underline{\Diamond } \boldsymbol{\mathit{w} } ^{\mathit{B}}$ is equivalent to averaging the view-specific evidences $\boldsymbol{\mathit{e}} ^{A\underline{\Diamond } B}= \frac{1}{2} (\boldsymbol{\mathit{e}} ^{\mathit{A}} + \boldsymbol{\mathit{e}} ^{\mathit{B}} )$.

Based on this proposition, in the multi-view fusion stage, we establish a simple and effective average pooling fusion layer, 
$g(\cdot)$, to realize conflictive opinion aggregation. 

\textbf{Proposition 2} For the conflictive opinion aggregation, after aggregating a new opinion into the original opinion, if the uncertain mass of the new opinion is smaller than the original uncertain mass, the uncertain mass of the aggregated opinion would be smaller than the original one; conversely, it would be larger.

An important characteristic of most existing trust multi-view learning methods \cite{han2021trusted,jung2022uncertainty,Liu_Yue_Chen_Denoeux_2022,xu2022uncertainty,liu2023safe,zhang2023provable} is ``After integrating another opinion into the original opinion, the obtained uncertainty mass will be reduced." We argue that this is unreasonable since: 1) when integrating a reliable perspective, the fusion process should ideally reduce the overall uncertainty; 2) when incorporating an unreliable or conflicting perspective, the fusion should increase the uncertainty. Furthermore, existing methods often overlook the possibility of conflicts between opinions gathered from different views. These conflicts may arise due to misaligned data or variations in the model's performance across different views.

To illustrate this issue, let's consider the scenario of two observers, A and B, observing colored balls drawn from a box. The observers could be seem as sensors to collect multi-view data. Different kinds of observers (normal or color blind) can produce conflictive multi-view data. The balls can be one of four colors: black, white, red, or green. Observer B is color blind, specifically having difficulty distinguishing between red and green balls while being able to differentiate between other color combinations. On the other hand, Observer A has perfect color vision and can usually identify the correct color when a ball is selected. Consequently, when a red ball is chosen, observer A typically identifies it as red, while observer B may perceive it as green. This disagreement between A and B leads to conflicting opinions regarding the same object.

Assuming that it is initially unknown whether one of the observers is color blind, their opinions are considered equally reliable. However, the existing fusion methods would erroneously reduce the uncertainty after combining the opinions of both observers. But in this case, we should treat the information from each perspective equally, given the possibility of conflicting opinions, and not automatically reduce the uncertainty. In summary, existing fusion methods should take into account the potential conflicts between opinions collected from different views and treat each perspective's information equally, rather than assuming a reduction in uncertainty based solely on the fusion process.

\section{Experiments}
In this section, we evaluate ECML on 6 real-world multi-view datasets. Furthermore, we also analyze the conflictive degree and uncertainty on conflictive multi-view data.

\subsection{Experimental Setup}

\subsubsection{Datasets.}

\textbf{HandWritten}\footnote{https://archive.ics.uci.edu/ml/datasets/Multiple+Features} comprises 2000 instances of handwritten numerals ranging from '0' to '9', with 200 patterns per class. It is represented using six feature sets. \textbf{CUB}\footnote{http://www.vision.caltech.edu/visipedia/CUB-200.html}consists of 11788 instances associated with text descriptions of 200 different categories of birds. In this study, we focus on the first 10 categories and extract image features using GoogleNet and corresponding text features using doc2vec. \textbf{HMDB}\footnote{https://serre-lab.clps.brown.edu/resource/hmdb-a-large-human-motion-database} is a large-scale human action recognition dataset containing 6718 instances from 51 action categories. We extract the HOG and MBH features as multiple views for this dataset. \textbf{Scene15}\footnote{https://doi.org/10.6084/m9.figshare.7007177.v1} includes 4485 images from 15 indoor and outdoor scene categories. We extract three types of features GIST, PHOG, and LBP. \textbf{Caltech101}\footnote{http://www.vision.caltech.edu/Image Datasets/Caltech101} comprises 8677 images from 101 classes. We select the first 10 categories and extract two deep features (views) using DECAF and VGG19 models. \textbf{PIE}\footnote{http://www.cs.cmu.edu/afs/cs/project/PIE/MultiPie/Multi-Pie/Home.html} contains 680 instances belonging to 68 classes. We extract intensity, LBP, and Gabor as 3 views. Table \ref{dataset} summarizes a summary of the datasets.

\begin{table}[t]
\centering
\small
\begin{tabular}{|l|c|c|c|c|}
\hline
\multicolumn{1}{|c|}{Dataset} & Size & $\mathit{K}$ & Dimensionality  \\ 
\hline
\hline
\multicolumn{1}{|c|}{HandWritten} & 2000 & 10 & 240/76/216/47/64/6 \\ 
\hline
\multicolumn{1}{|c|}{CUB} & 11788 & 10 & 1024/300 \\ 
\hline
\multicolumn{1}{|c|}{HMDB} & 6718 & 51 & 1000/1000 \\ 
\hline
\multicolumn{1}{|c|}{Scene15} & 4485 & 15 & 20/59/40 \\ 
\hline
\multicolumn{1}{|c|}{Caltech101} & 8677 & 101 & 4096/4096 \\ 
\hline
\multicolumn{1}{|c|}{PIE} & 680 & 68 & 484/256/279 \\ 
\hline
\end{tabular}
\centering \caption{ \label{dataset} Dataset summary.}
\end{table}

\begin{table*}[htb]
\centering
\begin{tabular*}{0.9\textwidth}{@{\extracolsep{\fill}}cccccccc}
\hline
Data        & DCCAE          & CPM-Nets       & DUA-Nets       & TMC            & TMDL-OA      & Ours          & $\bigtriangleup \%$  \\ 
\hline
HandWritten & 95.45$\pm$0.35 & 94.55$\pm$1.36 & 98.10$\pm$0.32 & 98.51$\pm$0.13 & \underline{99.25$\pm$0.45} & $\mathbf{99.40 \pm 0.00}$ & 0.15 \\

CUB         & 85.39$\pm$1.36 & 89.32$\pm$0.38 & 80.13$\pm$1.67 & 90.57$\pm$2.96 & \underline{95.43$\pm$0.20} & $\mathbf{98.50 \pm 2.75}$ & 3.21 \\

HMDB        & 49.12$\pm$1.07 & 63.32$\pm$0.43 & 62.73$\pm$0.23 & 65.17$\pm$2.42 & \underline{88.20$\pm$0.58} & $\mathbf{90.84 \pm 1.86}$ & 2.99 \\

Scene15     & 55.03$\pm$0.34 & 67.29$\pm$1.01 & 68.23$\pm$0.11 & 67.71$\pm$0.30 & \underline{75.57$\pm$0.02} & $\mathbf{76.19 \pm 0.12}$ & 0.82 \\

Caltech101  & 89.56$\pm$0.41 & 90.35$\pm$2.12 & 93.43$\pm$0.34 & 92.80$\pm$0.50 & \underline{94.63$\pm$0.04} & $\mathbf{95.36 \pm 0.38}$ & 0.77 \\

PIE         & 81.96$\pm$1.04 & 88.53$\pm$1.23 & 90.56$\pm$0.47 & 91.85$\pm$0.23 & \underline{92.33$\pm$0.36} & $\mathbf{94.71 \pm 0.02}$ & 2.57 \\ 
\hline
\end{tabular*}
\caption{ \label{normal_acc} Accuracy (\%) on normal test sets. The best and the second best results are highlighted by boldface and underlined respectively. $\bigtriangleup \%$ denotes the performance improvement of ECML over the best baseline.}
\end{table*}

\begin{table*}[htb]
\centering
\begin{tabular*}{0.9\textwidth}{@{\extracolsep{\fill}}cccccccc}
\hline
Data        & DCCAE          & CPM-Nets       & DUA-Nets       & TMC            & TMDL-OA      & Ours          & $\bigtriangleup \%$  \\ 
\hline
HandWritten & 82.85$\pm$0.38 & 83.34$\pm$1.07 & 87.16$\pm$0.34 & 92.76$\pm$0.15 & \underline{93.05$\pm$0.05} & $\mathbf{94.40 \pm 0.05}$ & 1.45\\

CUB         & 63.57$\pm$1.28 & 68.82$\pm$0.17 & 60.53$\pm$1.17 & 73.37$\pm$2.16 & \underline{74.43$\pm$0.26} & $\mathbf{76.50 \pm 1.15}$ & 2.78\\

HMDB        & 29.62$\pm$1.79 & 42.62$\pm$1.43 & 43.53$\pm$0.28 & 47.17$\pm$0.15 & \underline{67.62$\pm$0.28} & $\mathbf{70.84 \pm 1.19}$ & 4.76\\

Scene15     & 25.97$\pm$2.86 & 29.63$\pm$1.12 & 26.18$\pm$1.31 & 42.27$\pm$1.61 & \underline{48.42$\pm$1.02} & $\mathbf{56.97 \pm 0.52}$ & 17.66\\

Caltech101  & 60.90$\pm$2.32 & 66.54$\pm$2.89 & 75.19$\pm$2.34 & 90.16$\pm$2.50 & \underline{90.63$\pm$2.05} & $\mathbf{92.36 \pm 1.48}$ & 1.91\\

PIE         & 26.89$\pm$1.10 & 53.19$\pm$1.17 & 56.45$\pm$1.75 & 61.65$\pm$1.03 & \underline{68.16$\pm$0.34} & $\mathbf{84.00 \pm 0.14}$ & 23.24\\
\hline
\end{tabular*}
\caption{ \label{conflict_acc} Accuracy (\%) on conflictive test sets.}
\end{table*}

\subsubsection{Compared Methods.}
The baselines based on feature fusion include: (1) \textbf{DCCAE} (Deep Canonically Correlated AutoEncoders) \cite{10.5555/3045118.3045234} is the classical method, which employs autoencoders to seek a common representation. (2) \textbf{CPM-Nets} (Cross Partial Multi-view Networks) \cite{9258396} is a SOTA multi-view feature fusion method, which focuses on learning a versatile representation to handle complex correlations among different views. (3) \textbf{DUA-Nets} (Dynamic Uncertainty-Aware Networks) \cite{Geng_Han_Zhang_Hu_2021} is an uncertainty-aware method, which utilizes reversal networks to integrate intrinsic information from different views into a unified representation. The baselines based on decision fusion include: (1) \textbf{TMC} (Trusted Multi-view Classification) \cite{han2021trusted} is the pioneer uncertainty-aware method, which addresses the uncertainty estimation problem and produces reliable classification results. (2) \textbf{TMDL-OA} (Trusted Multi-View Deep Learning with Opinion Aggregation) \cite{Liu_Yue_Chen_Denoeux_2022} is a SOTA multi-view decision fusion method, which is also based on the evidential DNN and proposes a consistency measure loss to achieve trustworthy learning results.


To create a test set with conflictive instances, we perform the following transformations: (1) For noise views, we introduce Gaussian noise with varying levels of standard deviations $\sigma$ to a partial percentage of the test instances. (2) For unaligned views, we select a portion of the instances and modify the information of a random view, causing the label corresponding to that view to be misaligned with the true label of the instance. We conduct 10 runs for each method and report the mean values and standard deviations.

\subsection{Experiment Results}

\subsubsection{Performance Comparison.}

Tables \ref{normal_acc} and \ref{conflict_acc} show the classification performance on normal and conflictive test sets, respectively. We obtain that: (1) Even on normal test sets, ECML  outperforms all the other baselines. For instance, on the HMDB dataset, ECML achieves an accuracy improvement of approximately $2.64\%$ compared to the second-best (TMDL-OA) model. The reason would be attributed to the incorporation of consistency loss, which enhances the model's learning capability, as validated by the ablation study. (2) When evaluating on conflictive test sets, the accuracy of all the compared methods notably decreases. Nonetheless, thanks to the conflictive opinion aggregation, ECML exhibits an awareness of view-specific conflicts, leading to impressive results across all datasets. This highlights the effectiveness of ECML for both normal and conflictive multi-view data.


\subsubsection{Conflictive Degree Visualization}
\begin{figure}[t]
    \centering
    \includegraphics[width=0.9\columnwidth]{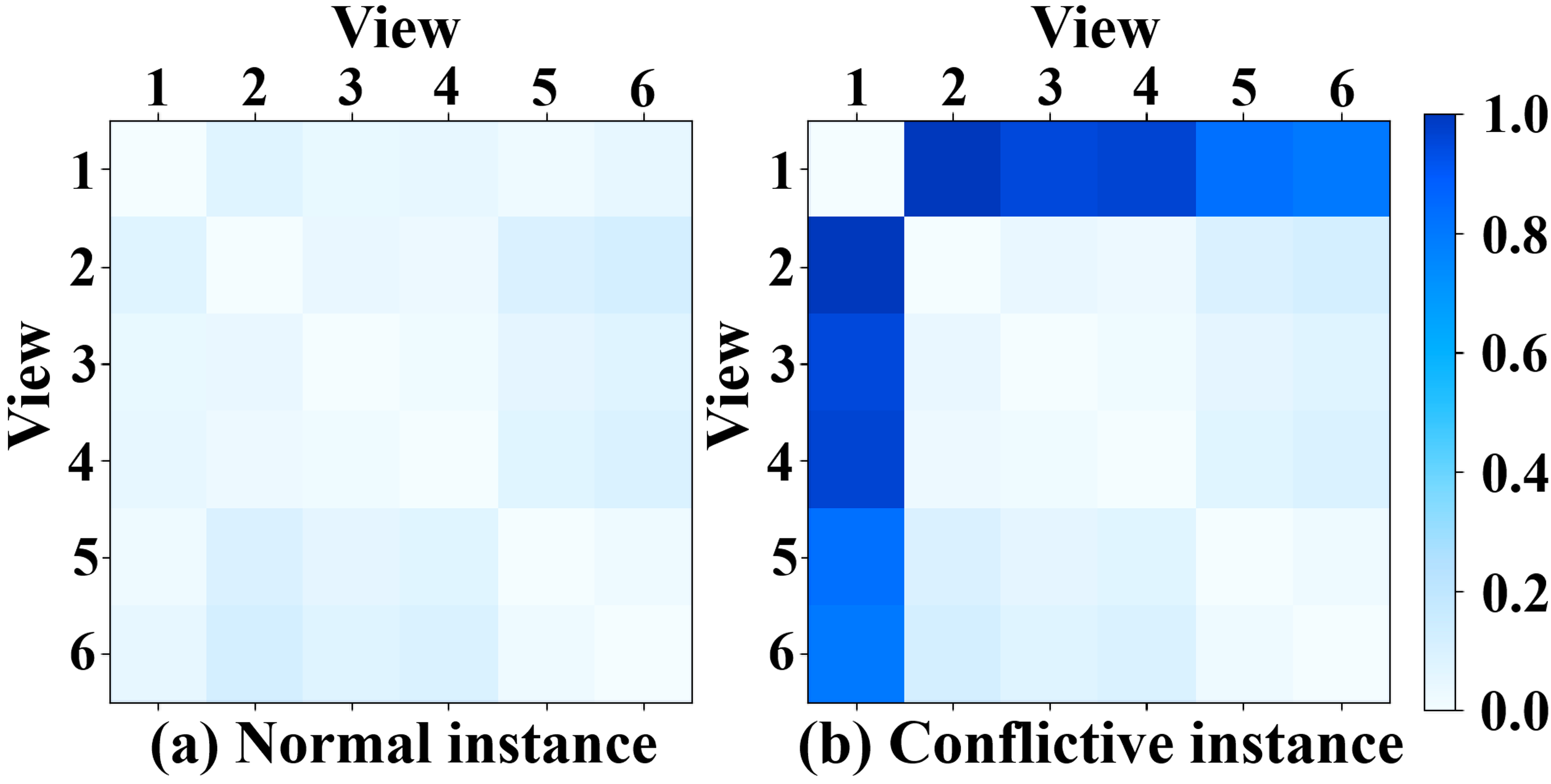}
    \caption{Conflictive degree visualization.}
    \label{fig:conflict}
\end{figure}

Fig. \ref{fig:conflict} shows the conflictive degree on the HandWritten dataset with six views. The left and right parts show the conflictive degree of normal and conflictive instances, respectively. To create conflicts, we modify the content from the first view, causing unaligned content from the other views. The results clearly demonstrate that ECML can effectively capture and quantify conflictive degrees between views. This finding further validates the reliability of ECML.

\subsubsection{Uncertainty Estimation}

\begin{figure}[t]
    \centering
    \subfigure[$\sigma=0.1$]{
        \centering
        \includegraphics[width=0.45\columnwidth]{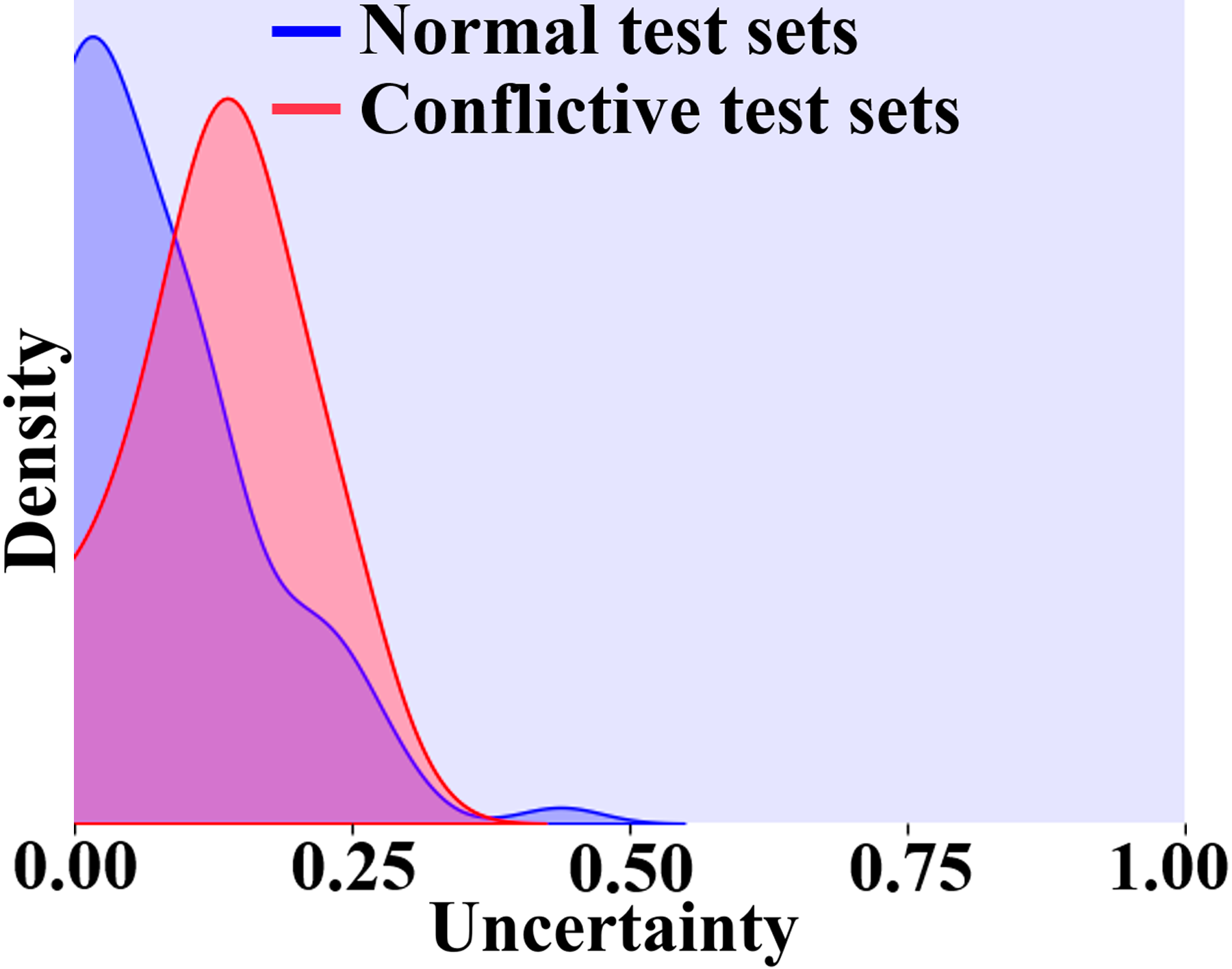}
    }
    \hfill 
    \subfigure[$\sigma=1$]{
        \centering
        \includegraphics[width=0.45\columnwidth]{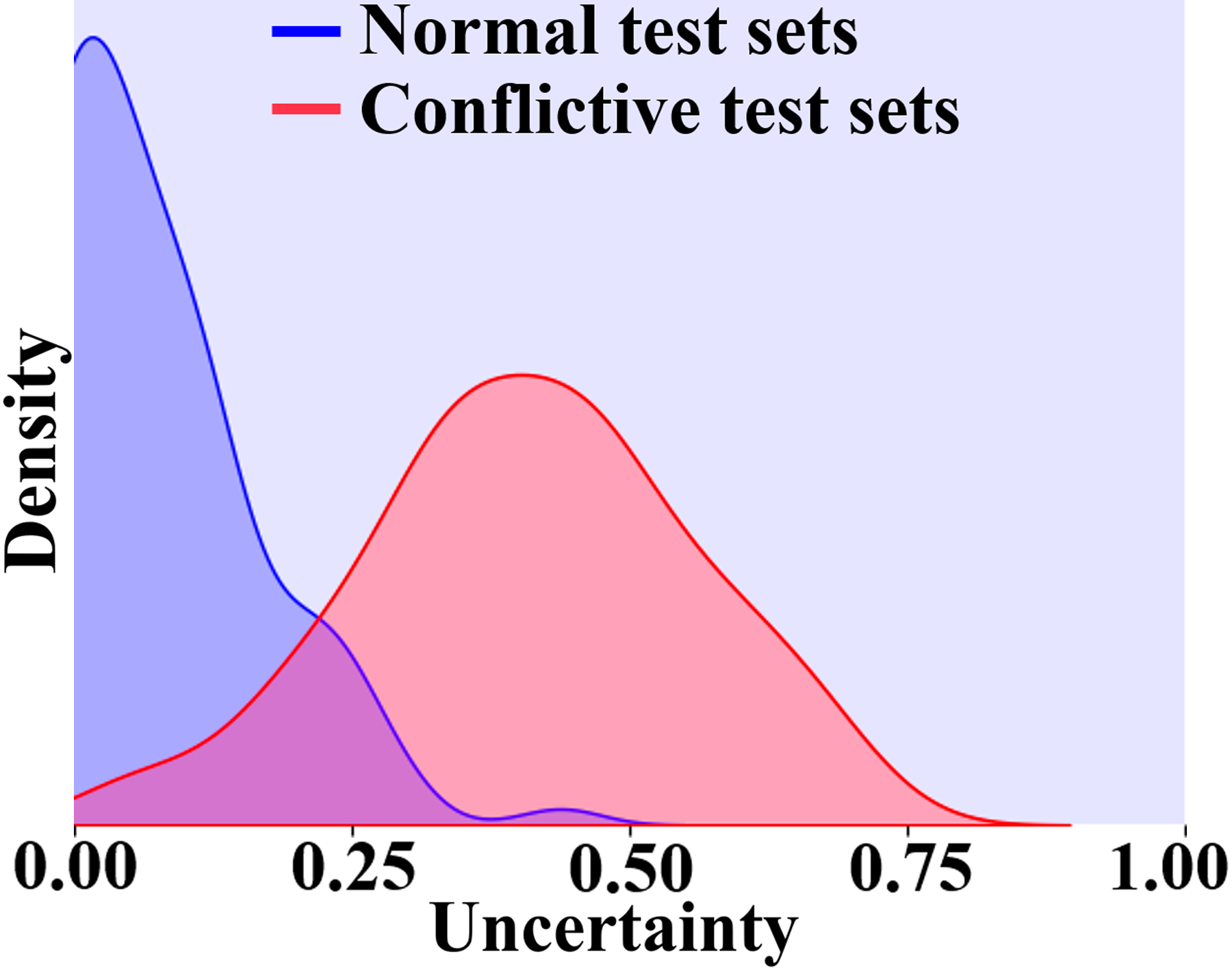}
    }
    \newline
    \subfigure[$\sigma=5$]{
        \centering
        \includegraphics[width=0.45\columnwidth]{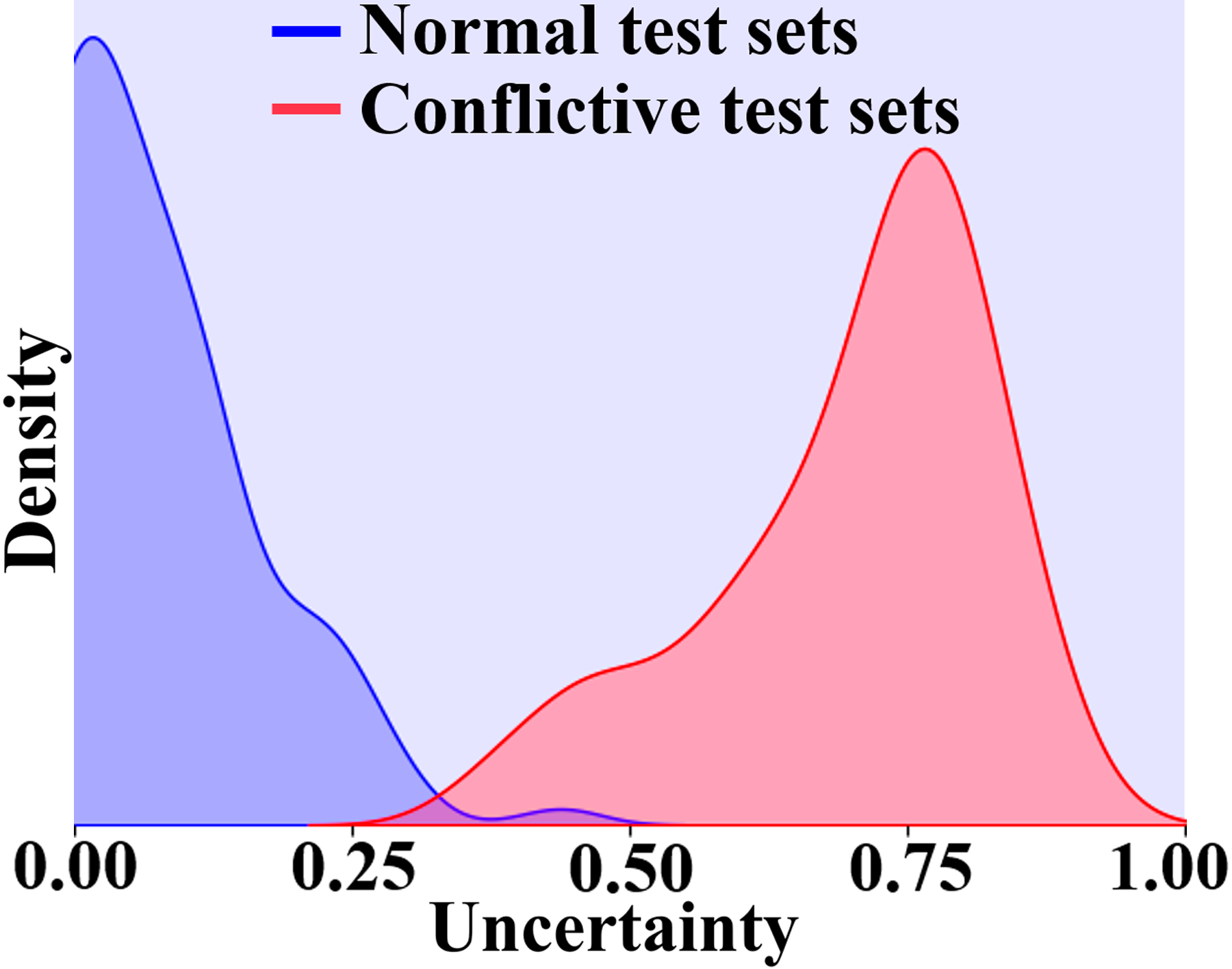}
    }
    \hfill 
    \subfigure[$\sigma=10$]{
        \centering
        \includegraphics[width=0.45\columnwidth]{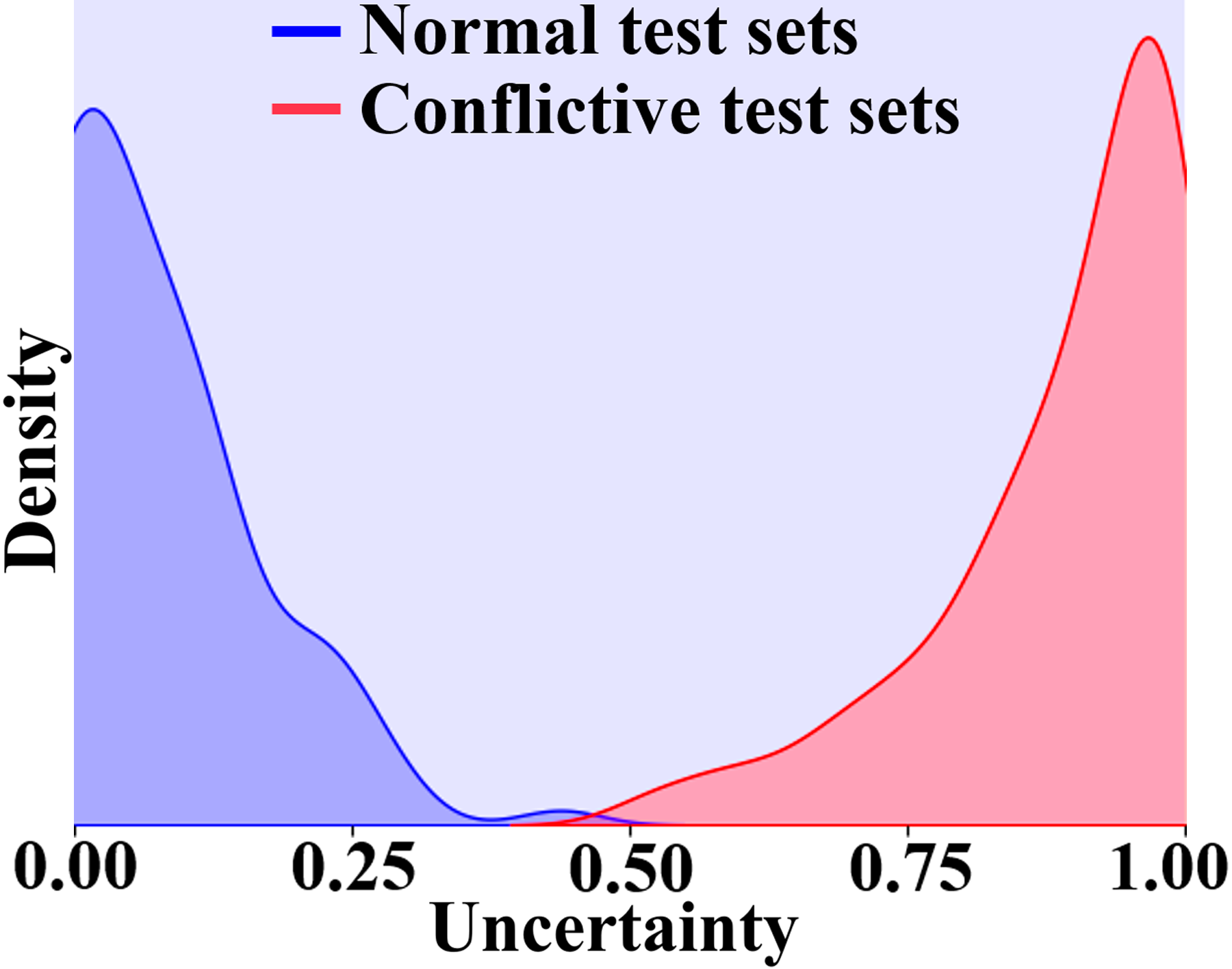}
    }
    \caption{\label{distribution} Density of uncertainty.}
\end{figure}

To further evaluate the estimated uncertainty, we visualize the distribution of normal and conflictive test sets on the CUB dataset. To construct conflictive test sets, we introduce Gaussian noise with standard deviation $\sigma = 0.1, 1, 5, 10$ to $50\%$ of the test instances. The experimental results are presented in Fig. \ref{distribution}. The results reveal that, when the noise intensity is low ($\sigma = 0.1$), the distribution curve of the conflictive instances closely aligns with that of the normal instances. However, as the noise intensity increases, the uncertainty of the conflictive instances also increases. This finding indicates that the estimated uncertainty is correlated with the quality of the instances, thereby validating the capability of our method in uncertainty estimation. 

\section{Conclusion}

In this paper, we proposed an Evidential Conflictive Multi-view Learning (ECML) method for the RCML problem. ECML tries to form view-specific opinions consisting of belief mass vector and decision reliability. It further aggregates conflictive opinions by a simple and effective average pooling layer. We theoretically proved it can exactly model the relation of multi-view common and view-specific reliabilities. Furthermore, we also extended our method by minimizing the degree of conflict between opinions to guarantee the consistency of results between different opinions. Experimental results on six real-world datasets confirmed the effectiveness of ECML. 

\section{Acknowledgments}
This research was supported by the National Natural Science Foundation of China (Grant Nos. 62133012, 61936006, 62103314, 62302370, 62073255, 62303366), the Key Research and Development Program of Shanxi (Program No. 2020ZDLGY04-07), Innovation Capability Support Program of Shanxi (Program No. 2021TD-05), Natural Science Basic Research Program of Shaanxi under Grant No.2023-JC-QN-0648, the Open Project of Anhui Provincial Key Laboratory of Multimodal Cognitive Computation, Anhui University, No. MMC202105.

\bibliography{aaai24}

\end{document}